\documentclass[twoside,11pt]{article}

\usepackage{graphicx}
\usepackage{url}
\usepackage{stfloats}
\usepackage{amssymb,amsmath}
\usepackage{amsfonts}
\usepackage{array}
\usepackage{cases}

\usepackage{jmlr2e}
\usepackage[multiple]{footmisc}

\firstpageno{1}

\begin{document}

\title{Memory-Efficient Topic Modeling}

\author{\name Jia Zeng \email j.zeng@ieee.org \\
       \addr School of Computer Science and Technology \\
       Soochow University\\
       Suzhou 215006, China
       \AND
       \name Zhi-Qiang Liu and Xiao-Qin Cao \\
       \addr School of Creative Media\\
       City University of Hong Kong\\
       Tat Chee Ave 83, Kowloon Tong, Hong Kong
       }

\editor{}

\maketitle

\begin{abstract}
As one of the simplest probabilistic topic modeling techniques,
latent Dirichlet allocation (LDA) has found many important applications in text mining,
computer vision and computational biology.
Recent training algorithms for LDA can be interpreted within a unified message passing framework.
However,
message passing requires storing previous messages with a large amount of memory space,
increasing linearly with the number of documents or the number of topics.
Therefore,
the high memory usage is often a major problem for topic modeling of massive corpora containing a large number of topics.
To reduce the space complexity,
we propose a novel algorithm without storing previous messages for training LDA:
tiny belief propagation (TBP).
The basic idea of TBP relates the message passing algorithms with the non-negative matrix factorization (NMF) algorithms,
which absorb the message updating into the message passing process,
and thus avoid storing previous messages.
Experimental results on four large data sets confirm that TBP performs comparably well or even better than
current state-of-the-art training algorithms for LDA but with a much less memory consumption.
TBP can do topic modeling when massive corpora cannot fit in the computer memory,
for example,
extracting thematic topics from $7$GB PUBMED corpora on a common desktop computer with $2$GB memory.
\end{abstract}


\section{Introduction}

Latent Dirichlet allocation (LDA)~\citep{Blei:03} is a three-layer hierarchical Bayesian model
for probabilistic topic modeling, computer vision and computational biology~\citep{Blei:12}.
The collections of documents can be represented as a document-word co-occurrence matrix,
where each element is the number of word count in the specific document.
Modeling each document as a mixture topics and each topic as a mixture of vocabulary words,
LDA assigns thematic labels to explain non-zero elements in the document-word matrix,
segmenting observed words into several thematic groups called topics.
From the joint probability of latent labels and observed words,
existing training algorithms of LDA approximately infers the posterior probability of topic labels given observed words,
and estimate multinomial parameters for document-specific topic proportions and topic distributions of vocabulary words.
The time and space complexity of these training algorithms depends on the number of non-zero ($NNZ$) elements in the matrix.

Probabilistic topic modeling for massive corpora has attracted intense interests recently.
This research line is motivated by increasingly common massive data sets,
such as online distributed texts, images and videos.
Extracting and analyzing the large number of topics from these massive data sets brings new challenges to current topic modeling algorithms,
particularly in computation time and memory requirement.
In this paper,
we focus on reducing the memory usage of topic modeling for massive corpora,
because the memory limitation prohibits running existing topic modeling algorithms.
For example,
when the document-word matrix has $NNZ=5\times10^8$,
existing training algorithms of LDA often requires allocating more than $12$GBytes memory including space for data and parameters.
Such a topic modeling task cannot be done on a common desktop computer with $2$GB memory even if we can tolerate the slow speed of topic modeling.

Because computing the exact posterior of LDA is intractable,
we must adopt approximate inference methods for training LDA.
Modern approximate posterior inference algorithms for LDA fall broadly into three categories:
variational Bayes (VB)~\citep{Blei:03},
collapsed Gibbs sampling (GS)~\citep{Griffiths:04},
and loopy belief propagation (BP)~\citep{Zeng:11}.
We may interpret these methods within a unified message passing framework~\citep{Bishop:book},
which infers the approximate marginal posterior distribution of the topic label for each word called {\em message}.
According to the expectation-maximization (EM) algorithm~\citep{Dempster:77},
the local inferred messages are used to estimate the best multinomial parameters in LDA based on the maximum-likelihood (ML) criterion.

VB is a variational message passing algorithm~\citep{Winn:05},
which infers the message from a factorizable variational distribution to be close in Kullback-Leibler (KL) divergence to the joint distribution.
The gap between variational and true joint distributions cause VB to use computationally expensive digamma functions,
introducing biases and slowness in the message updating and passing process~\citep{Asuncion:09,Zeng:11}.
GS is based on Markov chain Monte Carlo (MCMC) sampling process,
whose stationary distribution is the desired joint distribution.
GS usually updates its message using the sampled topic labels from previous messages,
which does not keep all uncertainties of previous messages.
In contrast,
BP directly updates and passes the entire messages without sampling,
and thus achieves a much higher topic modeling accuracy.
Till now,
BP is very competitive in both speed and accuracy for topic modeling~\citep{Zeng:11}.
Similar BP ideas have also been discussed as the zero-order approximation of the collapsed VB (CVB0)
algorithm within the mean-field framework~\citep{Asuncion:09,Asuncion:10}.

However,
the message passing techniques often require storing previous messages for updating and passing,
which leads to the high memory usage increasing linearly with the number of documents or the number of topics.
So,
to save the memory usage,
we propose a novel algorithm for training LDA: tiny belief propagation (TBP).
The basic idea of TBP is inspired by the multiplicative update rules of non-negative matrix factorization (NMF)~\citep{Lee:01},
which absorbs the message updating into passing process without storing previous messages.
Extensive experiments demonstrate that TBP enjoys a significantly less memory usage for topic modeling of massive data sets,
but achieves a comparable or even better topic modeling accuracy than VB, GS and BP.
Moreover,
the speed of TBP is very close to BP,
which is currently the fastest batch learning algorithm for topic modeling~\citep{Zeng:11}.
We also extend the proposed TBP using the block optimization framework~\citep{Yu:10}
to handle the case when data cannot fit in computer memory.
For example,
we extend TBP to extract $10$ topics from $7$GB PUBMED biomedical corpus using a desktop computer with $2$GB memory.

There have been two straight-forward machine learning strategies to process large-scale data sets:
online and parallel learning schemes.
On the one hand,
online topic modeling algorithms such as online VB (OVB)~\citep{Hoffman:10} read massive corpora as a data stream composed of multiple smaller mini-batches.
Loading each smaller mini-batch into memory,
OVB optimizes LDA within the online stochastic optimization framework,
theoretically converging to the batch VB's objective function.
But OVB still needs to store messages for each mini-batch.
When the size of mini-batch is large,
the space complexity of OVB is still higher than the batch training algorithm TBP.
In addition,
the best online topic modeling performance depends highly on several heuristic parameters including the mini-batch size.
On the other hand,
parallel topic modeling algorithms such as parallel GS (PGS)~\citep{Newman:09} use expensive parallel architectures with more physical memory.
Indeed,
PGS does not reduce the space complexity for training LDA,
but it distributes massive corpora into $P$ distributed computing units,
and thus requires only $1/P$ memory usage as GS.
By contrast,
the proposed TBP can reduce the space complexity for batch training LDA on a common desktop computer.
Notice that we may also develop much more efficient online and parallel topic modeling algorithms based on TBP in order for a significant speedup.

The rest paper is organized as follows.
Section~\ref{s2} compares VB, GS and BP for message passing,
and analyzes their space complexity for training LDA.
Section~\ref{s3} proposes the TBP algorithm to reduce the space complexity of BP,
and discusses TBP's relation with the multiplicative update rules of NMF.
Section~\ref{s4} shows extensive experiments on four real-world corpora.
Finally,
Section~\ref{s5} draws conclusions and envisions future work.

\section{The Message Passing Algorithms for Training LDA} \label{s2}

\begin{figure*}
\centering
\includegraphics[width=0.6\linewidth]{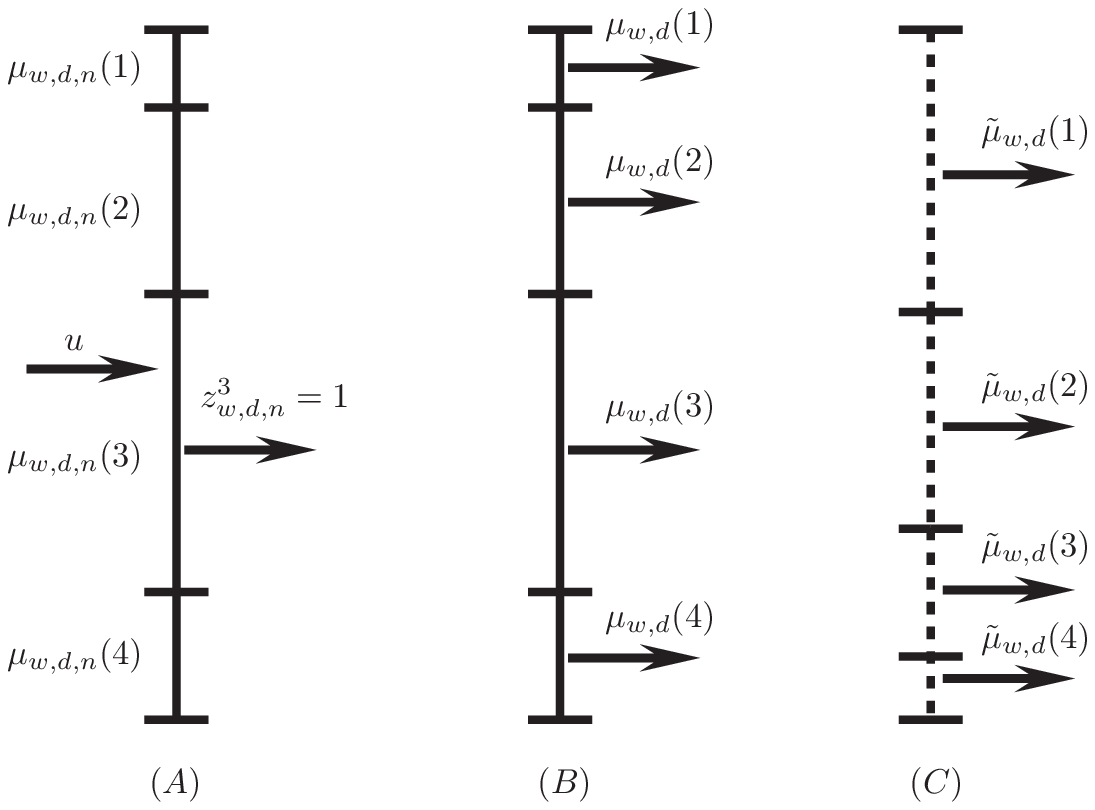}
\caption{Message passing for training LDA: (A) collapsed Gibbs sampling (GS), (B) loopy belief propagation (BP), and (C) variational Bayes (VB).}
\label{message}
\end{figure*}

LDA allocates a set of semantic topic labels,
$\mathbf{z} = \{z^k_{w,d}\}$,
to explain non-zero elements
in the document-word co-occurrence matrix $\mathbf{x}_{W \times D} = \{x_{w,d}\}$,
where $1 \le w \le W$ denotes the word index in the vocabulary,
$1 \le d \le D$ denotes the document index in the corpus,
and $1 \le k \le K$ denotes the topic index.
Usually,
the number of topics $K$ is provided by users.
The topic label satisfies $z^k_{w,d} = \{0,1\}, \sum_{k=1}^K z^k_{w,d} = 1$.
After inferring the topic labeling configuration over the document-word matrix,
LDA estimates two matrices of multinomial parameters:
topic distributions over the fixed vocabulary $\boldsymbol{\phi}_{W \times K} = \{\phi_{\cdot,k}\}$,
where $\theta_{\cdot, d}$ is a $K$-tuple vector and $\phi_{\cdot,k}$ is a $W$-tuple vector,
satisfying $\sum_k \theta_{k,d} = 1$ and $\sum_w \phi_{w,k} = 1$.
From a document-specific proportion $\theta_{\cdot,d}$,
LDA independently generates a topic label $z^k_{\cdot,d}=1$,
which further combines $\phi_{\cdot,k}$ to generate a word index $w$,
forming the total number of observed word counts $x_{w,d}$.
Both multinomial vectors $\theta_{\cdot,d}$ and $\phi_{\cdot,k}$ are generated by two Dirichlet distributions with hyperparameters $\alpha$ and $\beta$.
For simplicity,
we consider the smoothed LDA with fixed symmetric hyperparameters provided by users~\citep{Griffiths:04}.
To illustrate the generative process,
we refer the readers to the original three-layer graphical representation for LDA~\citep{Blei:03}
and the two-layer factor graph for the collapsed LDA~\cite{Zeng:11}.

Recently,
there have been three types of message passing algorithms for training LDA:
GS, BP and VB.
These message passing algorithms have space complexity as follows,
\begin{align}
\text{Total memory usage} = \text{data memory} + \text{message memory} + \text{parameter memory},
\end{align}
where the data memory is used to store the input document-word matrix $\mathbf{x}_{W \times D}$,
the message memory is allocated to store previous messages during passing,
and the parameter memory is used to store two output parameter matrices $\boldsymbol{\phi}_{W \times K}$ and $\boldsymbol{\theta}_{K \times D}$.
Because the input and output matrices of these algorithms are the same,
we focus on comparing the message memory consumption among these message passing algorithms.

\subsection{Collapsed Gibbs Sampling (GS)}

After integrating out the multinomial parameters $\{\phi,\theta\}$,
LDA becomes the collapsed LDA in the collapsed hidden variable space $\{\mathbf{z, \alpha, \beta}\}$.
GS~\citep{Griffiths:04} is a Markov Chain Monte Carlo (MCMC) sampling technique to infer the marginal distribution or {\em message},
$\mu_{w,d,n}(k) = p(z^k_{w,d,n}=1)$,
where $1 \le n \le x_{w,d}$ is the word token index.
The message update equation is
\begin{align} \label{GS}
\mu_{w,d,n}(k) \propto \frac{\mathbf{z}^k_{\cdot,d,-n} + \alpha}{\sum_k [\mathbf{z}^k_{\cdot,d,-n} + \alpha]} \times
\frac{\mathbf{z}^k_{w,\cdot,-n} + \beta}{\sum_w [\mathbf{z}^k_{w,\cdot,-n} + \beta]},
\end{align}
where $\mathbf{z}^k_{\cdot,d,-n} = \sum_{w} z^k_{w,d,-n}$,
$\mathbf{z}^k_{w,\cdot,-n} = \sum_{d} z^k_{w,d,-n}$,
and the notation $-n$ denotes excluding the current topic label $z^k_{w,d,n}$.
After normalizing the message $\sum_{k} \mu_{w,d,n}(k) = 1$,
GS draws a random number $u \sim \text{Uniform}[0,1]$ and checks which topic segment will be hit as shown in Fig.~\ref{message}A,
where $K = 4$ for example.
If the topic index $k=3$ is hit,
then we assign $z^3_{w,d,n} = 1$.
The sampled topic label will be used immediately to estimate the message for the next word token.
If we view the sampled topic labels as {\em particles},
GS can be interpreted as a special case of non-parametric belief propagation~\citep{Sudderth:03},
in which only particles rather than complete messages are updated and passed at each iteration.
Eq.~\eqref{GS} sweeps all word tokens for $1 \le t \le T$ training iterations until the convergence criterion is satisfied.
To exclude the current topic label $z^k_{w,d,n}$ in Eq.~\eqref{GS},
we need to store all topic labels,
$z^k_{w,d,n}=1, \forall w,d,n$,
in memory for message passing.
In a common $32$-bit desktop computer,
GS generally uses the integer type ($4$ bytes) for each topic label,
so the approximate message memory in bytes can be estimated by
\begin{align} \label{GSM}
\text{GS} = 4 \times \sum_{w,d} x_{w,d},
\end{align}
where $\sum_{w,d} x_{w,d}$ is the total number of word tokens in the document-word matrix.
For example,
$7$GB PUBMED corpus has $737,869,083$ word tokens,
occupying around $2.75$GB message memory according to Eq.~\eqref{GSM}.

Based on inferred topic configuration $z^k_{w,d,n}$ over word tokens,
the multinomial parameters can be estimated as follows,
\begin{gather}
\phi_{w,k} = \frac{\mathbf{z}^k_{w,\cdot,\cdot} + \beta}{\sum_{w} [\mathbf{z}^k_{w,\cdot,\cdot} + \beta]}, \\
\theta_{k,d} = \frac{\mathbf{z}^k_{\cdot,d,\cdot} + \alpha}{\sum_k [\mathbf{z}^k_{\cdot,d,\cdot} + \alpha]}.
\end{gather}
These equations look similar to Eq.~\eqref{GS} except including the current topic label $z^k_{w,d,n}$
in both numerator and denominator.

\subsection{Loopy Belief Propagation (BP)} \label{s2.2}

Similar to GS,
BP~\citep{Zeng:11} performs in the collapsed hidden variable space of LDA called collapsed LDA.
The basic idea is to integrate out the multinomial parameters $\{\theta,\phi\}$,
and infer the marginal posterior probability in the collapsed space $\{\mathbf{z},\alpha,\beta\}$.
The collapsed LDA can be represented by a factor graph,
which facilitates the BP algorithm for approximate inference and parameter estimation.
Unlike GS,
BP infers messages,
$\mu_{w,d}(k) = p(z^k_{w,d}=1)$,
without sampling in order to keep all uncertainties of messages.
The message update equation is
\begin{align} \label{BP}
\mu_{w,d}(k)\propto\frac{\boldsymbol{\mu}_{-w,d}(k) + \alpha}{\sum_k[\boldsymbol{\mu}_{-w,d}(k) + \alpha]} \times
\frac{\boldsymbol{\mu}_{w,-d}(k) + \beta}{\sum_w[\boldsymbol{\mu}_{w,-d}(k) + \beta]},
\end{align}
where $\boldsymbol{\mu}_{-w,d}(k) = \sum_{-w} x_{-w,d}\mu_{-w,d}(k)$ and
$\boldsymbol{\mu}_{w,-d}(k) = \sum_{-d} x_{w,-d}\mu_{w,-d}(k)$.
The notation $-w$ and $-d$ denote all word indices except $w$ and all document indices except $d$.
After normalizing $\sum_k \mu_{w,d}(k) = 1$,
BP updates other messages iteratively.
Fig.~\ref{message}B illustrates the message passing in BP when $K=4$,
slightly different from GS in Fig.~\ref{message}A.
Eq.~\eqref{BP} differs from Eq.~\eqref{GS} in two aspects.
First,
BP infers messages based on word indices rather than word tokens.
Second,
BP updates and passes complete messages without sampling.
In this sense,
BP can be viewed as a {\em soft} version of GS.
Obviously,
such differences give Eq.~\eqref{BP} two advantages over Eq.~\eqref{GS}.
First,
it keeps all uncertainties of messages for high topic modeling accuracy.
Second,
it scans a total of $NNZ$ word indices for message passing,
which is significantly less than the total number of word tokens $\sum_{w,d} x_{w,d}$ in $\mathbf{x}$.
So,
BP is often faster than GS by scanning a significantly less number of elements ($NNZ \ll \sum_{w,d}x_{w,d}$) at each training iteration~\citep{Zeng:11}.
Eq.~\eqref{BP} scans $NNZ$ in the document-word matrix for $1 \le t \le T$ training iterations until the convergence criterion is satisfied.

However,
BP has a higher space complexity than GS.
Because BP excludes the current message $\mu_{w,d}(k)$ in message update~\eqref{BP},
it requires storing all $K$-tuple messages.
In the widely-used $32$-bit desktop computer,
we generally use the double type ($8$ bytes) to store all messages with the memory occupancy in bytes,
\begin{align} \label{BPM}
\text{BP} = 8 \times K \times NNZ,
\end{align}
which increases linearly with the number of topics $K$.
For example,
$7$GB PUBMED corpus has $NNZ=483,450,157$.
When $K=10$,
BP needs around $36$GB for message passing.
Notice that when $K$ is large,
Eq.~\eqref{BPM} is significantly higher than Eq.~\eqref{GSM}.

Based on the normalized messages,
the multinomial parameters can be estimated by
\begin{gather}
\label{thetad}
\phi_{w,k} = \frac{\boldsymbol{\mu}_{w,\cdot}(k) + \beta}{\sum_w [\boldsymbol{\mu}_{w,\cdot}(k) + \beta]}, \\
\theta_{k,d} = \frac{\boldsymbol{\mu}_{\cdot,d}(k) + \alpha}{\sum_k [\boldsymbol{\mu}_{\cdot,d}(k) + \alpha]}.
\label{phiw}
\end{gather}
These equations look similar to Eq.~\eqref{BP} except including the current message $\mu_{w,d}(k)$ in both numerator and denominator.

\subsection{Variational Bayes (VB)}

Unlike BP in the collapsed space,
VB~\citep{Blei:03,Winn:05} passes variational messages,
$\tilde{\mu}_{w,d}(k) = \tilde{p}(z^k_{w,d}=1)$,
derived from the approximate variational distribution $\tilde{p}$ to the true joint distribution $p$
by minimizing the KL divergence, $KL(\tilde{p}||p)$.
The variational message update equation is
\begin{align} \label{VB}
\tilde{\mu}_{w,d}(k) \propto \frac{\exp[\Psi(\tilde{\boldsymbol{\mu}}_{\cdot,d}(k) + \alpha)]}
{\exp[\Psi(\sum_k [\tilde{\boldsymbol{\mu}}_{\cdot,d}(k) + \alpha])]} \times
\frac{\tilde{\boldsymbol{\mu}}_{w,\cdot}(k) + \beta}{\sum_w [\tilde{\boldsymbol{\mu}}_{w,\cdot}(k) + \beta]},
\end{align}
where $\tilde{\boldsymbol{\mu}}_{\cdot,d}(k) = \sum_{w} x_{w,d}\tilde{\mu}_{w,d}(k)$,
$\tilde{\boldsymbol{\mu}}_{w,\cdot}(k) = \sum_{d} x_{w,d}\tilde{\mu}_{w,d}(k)$,
and the notation $\exp$ and $\Psi$ are exponential and digamma functions, respectively.
After normalizing the variational message $\sum_k \tilde{\mu}_{w,d}(k) = 1$,
VB passes this message to update other messages.
There are two major differences between Eq.~\eqref{VB} and Eq.~\eqref{BP}.
First,
Eq.~\eqref{VB} involves computationally expensive digamma functions.
Second,
it include the current variational message $\tilde{\mu}_{w,d}$ in the update equation.
The digamma function significantly slows down VB,
and also introduces bias in message passing~\citep{Asuncion:09,Zeng:11}.
Fig.~\ref{message}C shows the variational message passing in VB,
where the dashed line illustrates that the variational message is derived from the variational distribution.
Because VB also stores the variational messages for updating and passing,
its space complexity is the same as BP in Eq.~\eqref{BPM}.
Based on the normalized variational messages,
VB estimates the multinomial parameters as
\begin{gather}
\phi_{w,k} = \frac{\tilde{\boldsymbol{\mu}}_{w,\cdot}(k) + \beta}{\sum_w [\tilde{\boldsymbol{\mu}}_{w,\cdot}(k) + \beta]}, \\
\theta_{k,d} = \frac{\tilde{\boldsymbol{\mu}}_{\cdot,d}(k) + \alpha}{\sum_k [\tilde{\boldsymbol{\mu}}_{\cdot,d}(k) + \alpha]}.
\end{gather}
These equations are almost the same as Eqs.~\eqref{thetad} and \eqref{phiw} but using variational messages.

\subsection{Synchronous and Asynchronous Message Passing} \label{s2.4}

Message passing algorithms for LDA first randomly initialize messages,
and then pass messages according to two schedules:
the synchronous and the asynchronous update schedules~\citep{Tappen:03}.
The synchronous message passing schedule uses all messages at $t-1$ training iteration to update current messages at $t$ training iteration,
while the asynchronous schedule immediately uses the updated messages to update other remaining messages within the same $t$ training iteration.
Empirical results demonstrate that the asynchronous schedule is slightly more efficient than the synchronous schedule~\citep{Zeng:11} for topic modeling.
However,
the synchronous schedule is much easier to extend for parallel computation.

GS is naturally an asynchronous message passing algorithm.
The sampled topic label will immediately influence the topic sampling process at the next word token.
Both synchronous and asynchronous schedules of BP work equally well in terms of topic modeling accuracy,
but the asynchronous schedule converges slightly faster than the synchronous one~\citep{Elidan:06}.
VB is a synchronous variational message passing algorithm,
updating messages at iteration $t$ using messages at iteration $t-1$.

\section{Tiny Belief Propagation} \label{s3}

In this section,
we propose TBP to save the message memory and data memory usage of BP in section~\ref{s2.2}.
Generally,
the parameter memory of BP takes a relatively smaller space when the number of topics $K$ is small.
For example,
as far as $7$GB PUBMED data set is concerned ($D = 8,200,000$ and $W = 141043$),
when $K=10$,
the parameter $\boldsymbol{\theta}_{K \times D}$ occupies around $0.6$GB memory,
while the parameter $\boldsymbol{\phi}_{W \times K}$ occupies around $0.01$GB memory.
For simplicity,
we assume that the parameter memory is enough for topic modeling.

\subsection{Message Memory} \label{3.1}

The algorithmic contribution of TBP is to reduce the message memory of BP to almost zero during message passing process.
Combining Eqs.~\eqref{BP},~\eqref{thetad} and~\eqref{phiw} yields the approximate message update equation,
\begin{align} \label{TBP}
\mu_{w,d}(k) \propto \phi_{w,k} \times \theta_{k,d},
\end{align}
where the current message $\mu_{w,d}(k)$ is added in both numerator and denominator in Eq.~\eqref{BP}.
Notice that such an approximation does not distort the message update very much
because the message $\mu_{w,d}(k)$ is significantly smaller than the aggregate of other messages in both numerator and denominator.
Eq.~\eqref{TBP} has the following intuitive explanation.
If the $w$th word has a higher likelihood in the topic $k$
and the topic $k$ has a larger proportion in the $d$th document,
then the topic $k$ has a higher probability to be assigned to the element $x_{w,d}$,
i.e.,
$z^k_{w,d} = 1$.
The normalized message can be written as the matrix operation,
\begin{align} \label{normalize}
\mu_{w,d}(k) = \frac{\phi_{w,k}\theta_{k,d}}{(\boldsymbol{\phi}\boldsymbol{\theta})_{w,d}},
\end{align}
where $(\boldsymbol{\phi}\boldsymbol{\theta})_{w,d}$ is the element at $\{w,d\}$ after matrix multiplication $\boldsymbol{\phi}\boldsymbol{\theta}$.
Within the probabilistic framework,
LDA generates the word token at index $\{w,d\}$ using the likelihood $(\boldsymbol{\phi}\boldsymbol{\theta})_{w,d}$,
which satisfies $\sum_{w} (\boldsymbol{\phi}\boldsymbol{\theta})_{w,d} = 1$,
so that $\sum_{w,d} (\boldsymbol{\phi}\boldsymbol{\theta})_{w,d} = D$ is a constant.
Replacing the normalized messages by Eq.~\eqref{normalize},
we can re-write Eqs.~\eqref{thetad} and~\eqref{phiw} as
\begin{gather}
\label{phi}
\phi_{w,k} \leftarrow \frac{\sum_{d}x_{w,d}[\phi_{w,k}\theta_{k,d}/(\boldsymbol{\phi}\boldsymbol{\theta})_{w,d}] + \beta}
{\sum_{w,d}x_{w,d}[\phi_{w,k}\theta_{k,d}/(\boldsymbol{\phi}\boldsymbol{\theta})_{w,d}] + W\beta}, \\
\theta_{k,d} \leftarrow \frac{\sum_{w}x_{w,d}[\phi_{w,k}\theta_{k,d}/(\boldsymbol{\phi}\boldsymbol{\theta})_{w,d}] + \alpha}{\sum_{w}x_{w,d} + K\alpha},
\label{theta}
\end{gather}
where the denominators play normalization roles to constrain $\sum_k \theta_{k,d} = 1, \theta_{k,d} \ge 0$ and $\sum_w \phi_{w,k} = 1, \phi_{w,k} \ge 0$.
We absorb the message update equation into the parameter estimation in Eqs.~\eqref{phi} and~\eqref{theta},
so that we do not need to store the previous messages during message passing process.
We refer to these matrix update algorithm as TBP.

If we discard the hyperparameters $\alpha$ and $\beta$ in Eqs.~\eqref{theta} and~\eqref{phi},
we find that these matrix update equations look similar to the following multiplicative update rules in non-negative matrix factorization (NMF)~\citep{Lee:01},
\begin{gather}
\label{nmf1}
\phi_{w,k} \leftarrow \frac{\sum_{d}x_{w,d}[\phi_{w,k}\theta_{k,d}/(\boldsymbol{\phi}\boldsymbol{\theta})_{w,d}]}{\sum_{d}\theta_{k,d}}, \\
\theta_{k,d} \leftarrow \frac{\sum_{w}x_{w,d}[\phi_{w,k}\theta_{k,d}/(\boldsymbol{\phi}\boldsymbol{\theta})_{w,d}]}{\sum_{w}\phi_{w,k}},
\label{nmf2}
\end{gather}
where the objective of NMF is to minimize the following divergence,
\begin{align} \label{divergence}
D(\mathbf{x}||\boldsymbol{\phi}\boldsymbol{\theta}) = \sum_{w,d}\bigg(x_{w,d}\log\frac{x_{w,d}}{(\boldsymbol{\theta}\boldsymbol{\phi})_{w,d}} -
x_{w,d} + (\boldsymbol{\theta}\boldsymbol{\phi})_{w,d}\bigg),
\end{align}
under the constraints $\phi_{w,d} \ge 0$ and $\theta_{k,d} \ge 0$.
First,
Eqs.~\eqref{theta} and~\eqref{phi} are different from Eqs.~\eqref{nmf1} and~\eqref{nmf2} in denominators,
just because LDA additionally constrain the sum of multinomial parameters to be one.
Second,
as far as LDA is concerned,
because $\sum_{w,d}x_{w,d}$ and $\sum_{w,d} (\boldsymbol{\phi}\boldsymbol{\theta})_{w,d}$ are constants,
Eq.~\eqref{divergence} is proportional to the standard Kullback-Leibler (KL) divergence,
\begin{align} \label{KL}
D(\mathbf{x}||\boldsymbol{\phi}\boldsymbol{\theta}) \propto
KL(\mathbf{x}||\boldsymbol{\phi}\boldsymbol{\theta}) &\propto \sum_{w,d}x_{w,d}\log\frac{x_{w,d}}{(\boldsymbol{\phi}\boldsymbol{\theta})_{w,d}} \notag \\
&\propto \sum_{w,d} -x_{w,d}\log(\boldsymbol{\phi}\boldsymbol{\theta})_{w,d}.
\end{align}
In conclusion,
if we discard hyperparameters in Eqs.~\eqref{phi} and Eq.~\eqref{theta},
the proposed TBP algorithm becomes a special NMF algorithm:
\begin{gather}
\mathbf{x} \approx \boldsymbol{\phi}\boldsymbol{\theta}, \\
\label{KL1}
\min \bigg(\sum_{w,d} -x_{w,d}\log(\boldsymbol{\phi}\boldsymbol{\theta})_{w,d}\bigg), \forall x_{w,d} \ne 0, \\
\phi_{w,k} \ge 0, \sum_w \phi_{w,k} = 1, \theta_{k,d} \ge 0, \sum_k \theta_{k,d} = 1,
\end{gather}
where TBP focuses only on approximating non-zero elements $x_{w,d} \ne 0$ by $\boldsymbol{\phi}\boldsymbol{\theta}$ in terms of the KL divergence.
Notice that the hyperparameters play smoothing roles in avoiding zeros in the factorized matrices in Eqs.~\eqref{phi} and Eq.~\eqref{theta},
where zeros are major reasons for worse performance in predicting unseen words in the test set~\citep{Blei:03}.

Conventionally,
different training algorithms for LDA can be fairly compared by the perplexity metric~\citep{Blei:03,Asuncion:09,Hoffman:10},
\begin{align} \label{perplexity}
\text{Perplexity} &= \exp\Bigg\{-\frac{\sum_{w,d}
x_{w,d}\log(\boldsymbol{\phi}\boldsymbol{\theta})_{w,d}}
{\sum_{w,d} x_{w,d}}\Bigg\}, \notag \\
&\propto \sum_{w,d} -x_{w,d}\log(\boldsymbol{\phi}\boldsymbol{\theta})_{w,d}.
\end{align}
which has been previously interpreted as the geometric mean of the likelihood in the probabilistic framework.
Comparing~\eqref{perplexity} with~\eqref{KL},
we find that the perplexity metric can be also interpreted as a KL divergence between
the document-word matrix $\mathbf{x}$
and the multiplication of two factorized matrices $\boldsymbol{\phi}\boldsymbol{\theta}$.
Because the TBP algorithm directly minimizes the KL divergence~\eqref{KL1},
it often has a much lower predictive perplexity on unseen test data than both GS and VB algorithms in Section~\ref{s2} for better topic modeling accuracy.
This theoretical analysis has also been supported by extensive experiments in Section~\ref{s4}.

Indeed,
LDA is a full Bayesian counterpart of the probabilistic latent semantic analysis (PLSA)~\citep{Hofmann:01},
which is equivalent to the NMF algorithm with the KL divergence~\citep{Gaussier:05}.
Moreover,
the inference objective functions between LDA and PLSA are very similar,
and PLSA can be viewed a maximum-a-posteriori (MAP) estimated LDA model~\citep{Girolami:03}.
For example,
two recent studies~\citep{Asuncion:09,Zeng:11} find that the CVB0 and the simplified BP algorithms for training LDA resemble the EM algorithm for training PLSA.
Based on these previous works,
it is a natural step to connect the NMF algorithms with those message passing algorithms for training LDA.
More generally,
we speculate that such intrinsic relations also exist between finite mixture models such as LDA and latent factor models such as NMF~\citep{Gershman:12}.
As an example,
the NMF algorithm in theory has been recently justified to learn topic models such as LDA with a polynomial time~\citep{Arora:12}.
Notice that TBP and other NMF algorithms do not need to store previous messages within the message passing framework in Section~\ref{s2},
and thus save a lot of memory usage.

Based on~\eqref{phi} and~\eqref{theta},
we implement two types of TBP algorithms:
synchronous TBP (sTBP) and asynchronous TBP (aTBP),
similar to the synchronous and asynchronous message passing algorithms in Section~\ref{s2.4}.
Because the denominator of~\eqref{theta} is a constant,
it does not influence the normalized message~\eqref{normalize}.
So,
we consider only the unnormalized $\theta$ during the matrix factorization.
However,
the denominator of~\eqref{phi} depends on $k$,
so we use a $K$-tuple vector $\boldsymbol{\lambda}_K$ to store the denominator,
and use the unnormalized $\boldsymbol{\phi}$ during the matrix factorization.
The normalization can be easily performed by a simple division $(\phi_{w,k} + \beta)/(\lambda_k + W\beta)$.

\begin{figure*}
\centering
\includegraphics[width=1\linewidth]{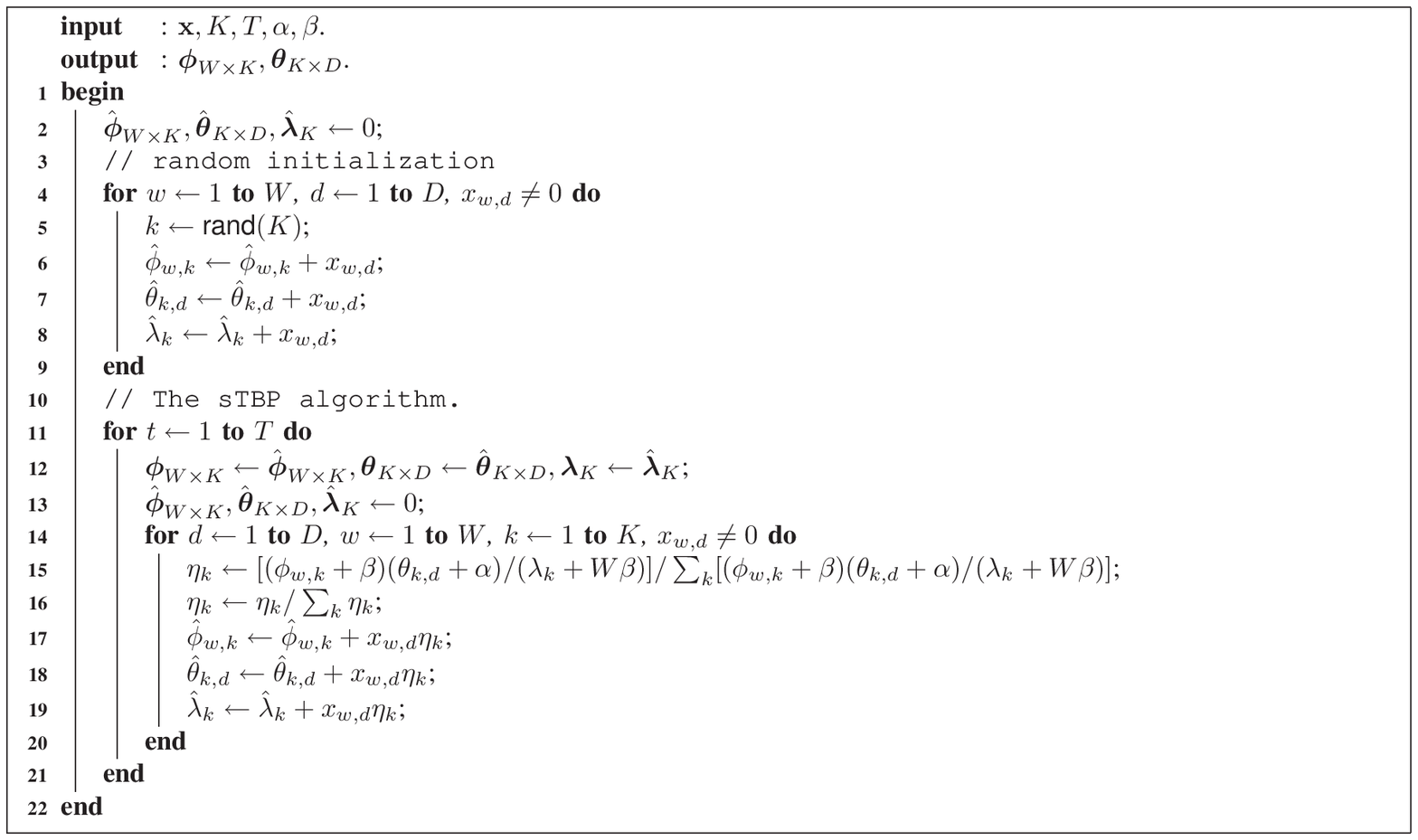}
\caption{The sTBP algorithm for LDA.}
\label{stbp}
\end{figure*}

\begin{figure*}
\centering
\includegraphics[width=1\linewidth]{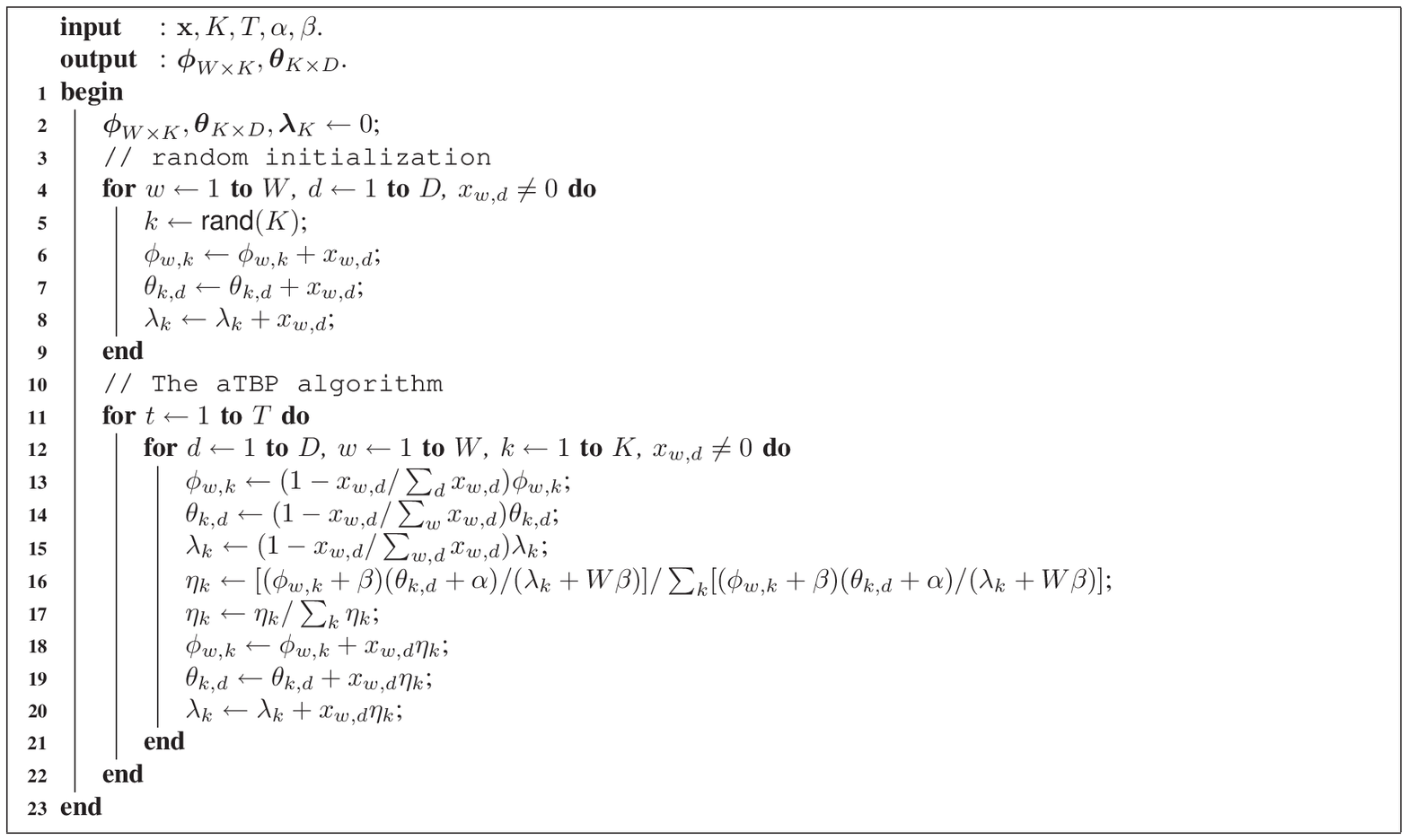}
\caption{The aTBP algorithm for LDA.}
\label{atbp}
\end{figure*}

Fig.~\ref{stbp} shows the synchronous TBP (sTBP) algorithm.
We use three temporary matrices $\hat{\boldsymbol{\phi}}, \hat{\boldsymbol{\theta}}, \hat{\boldsymbol{\lambda}}$ in Line $2$
to store numerators of~\eqref{phi},~\eqref{theta} and denominator of~\eqref{phi} for synchronization.
Form Line $4$ to $9$,
we randomly initialize $\hat{\boldsymbol{\phi}}, \hat{\boldsymbol{\theta}}, \hat{\boldsymbol{\lambda}}$ by $\verb+rand(K)+$,
which generates a random integer $k, 1 \le k \le K$.
At each training iteration $t, 1 \le t \le T$,
we copy the temporary matrices to $\boldsymbol{\phi}, \boldsymbol{\theta}, \boldsymbol{\lambda}$
and clear the temporary matrices to zeros from Line $12$ to $13$.
Then,
for each non-zero element in the document-word matrix,
we accumulate the numerators of~\eqref{phi},~\eqref{theta} and the denominator of~\eqref{phi} by the $K$-tuple message $\eta_k$
in the temporary matrices $\hat{\boldsymbol{\phi}}, \hat{\boldsymbol{\theta}}, \hat{\boldsymbol{\lambda}}$ from Line $15$ to $19$.
In the synchronous schedule,
the update of elements in the factorized matrices does not influence other elements within each iteration $t$.

Fig.~\ref{atbp} shows the asynchronous TBP (aTBP) algorithm.
Unlike the sTBP algorithm,
aTBP does not require temporary matrices $\hat{\boldsymbol{\phi}}, \hat{\boldsymbol{\theta}}, \hat{\boldsymbol{\lambda}}$.
After the random initialization from Line $4$ to $9$,
aTBP reduces the matrices $\boldsymbol{\phi}, \boldsymbol{\theta}, \boldsymbol{\lambda}$ in a certain proportion from Line $13$ to $15$,
which can be compensated by the updated $K$-tuple message $\eta_k$ from Line $17$ to $20$.
In the asynchronous schedule,
the change of elements in the factorized matrices $\boldsymbol{\phi}, \boldsymbol{\theta}$ will immediately influence the update of other elements.
In anticipation,
the asynchronous schedule is more efficient to pass the influence of the updated elements in matrices than the synchronous schedule.
The sTBP and aTBP algorithms will iterate until the convergence condition is satisfied or the maximum iteration $T$ is reached.

The time complexity of TBP is $\mathcal{O}(NNZ \times KT)$,
where $NNZ$ is the number of non-zero elements in the document-word matrix,
$K$ is the number of topics and $T$ is the number of training iterations.
sTBP has the space complexity $\mathcal{O}(3 \times NNZ + 2 \times KW + 2 \times KD)$,
but aTBP has the space complexity $\mathcal{O}(3 \times NNZ + KW + KD)$.
Generally,
we use $3 \times NNZ$ to store data in the memory including indices of non-zero elements in the document-word matrix,
and also use $KW + KD$ memory to store matrices $\boldsymbol{\phi}$ and $\boldsymbol{\theta}$.
Because sTBP uses additional matrices $\hat{\boldsymbol{\phi}}, \hat{\boldsymbol{\theta}}$ for synchronization,
it uses $2 \times KW + 2 \times KD$ for all matrices.

\subsection{Data Memory}

When the corpus data is larger than the computer memory,
traditional algorithms cannot train LDA due to the memory limitation.
We assume that the hard disk is large enough to store the corpus file.
Recently,
reading data from hard disk into memory as blocks is a promising method~\citep{Yu:10} to handle such problems.
We can extend the TBP algorithms in Figs.~\ref{stbp} and~\ref{atbp} to read the corpus file as blocks,
and optimize each block sequentially.
For example,
we can read each document in the corpus file at one time into memory
and perform the TBP algorithms to refine the matrices $\{\boldsymbol{\phi}, \boldsymbol{\theta}\}$.
After scanning all documents in the corpus data file,
TBP finishes one iteration of training in Figs.~\ref{stbp} and~\ref{atbp}.
Similarly,
we can also store the matrices $\{\boldsymbol{\phi}, \boldsymbol{\theta}\}$ in the file on the hard disk
when they are larger than computer memory.
In such cases,
TBP consumes almost no memory to do topic modeling.
Because loading data into memory requires additional time,
TBP running on files is around twice slower than that running on memory.
For example,
for the $7$GB PUBMED corpus and $K=10$,
aTBP requires $259.64$ seconds to scan the whole data file on the hard disk,
while it requires only $128.50$ seconds to scan the entire data in the memory at each training iteration.
Another choice is to extend TBP to the online learning~\citep{Hoffman:10},
which partitions the whole corpus file into mini-batches and optimizes each mini-batch after one look sequentially.
Although some online topic modeling algorithms like OVB can converge to the objective of corresponding batch topic modeling algorithm,
we find that the best topic modeling accuracy depends on several heuristic parameters including the mini-batch size~\citep{Hoffman:10}.
In contrast,
TBP is a batch learning algorithm that can handle large data memory with better topic modeling accuracy.
Reading block data from hard disk to memory can be also applied to both GS and VB algorithms for LDA.

\subsection{Relationship to Previous Algorithms}

The proposed TBP connects the training algorithm of LDA to the NMF algorithm with KL divergence.
The intrinsic relation between probabilistic topic models~\citep{Hofmann:01,Blei:03}
and NMF~\citep{Lee:01} have been extensively discussed in several previous
works~\citep{Buntine:02,Gaussier:05,Girolami:03,Wahabzada:11,Wahabzada:11a,Zeng:11}.
A more recent work shows that learning topic models by NMF has a polynomial time~\citep{Arora:12}.
Generally speaking,
learning topic models can be formulated within the message passing framework in Section~\ref{s2}
based on the generalized expectation-maximization (EM)~\citep{Dempster:77} algorithm.
The objective is to maximize the joint distribution of PLSA or LDA in two iterative steps.
At the E-step,
we approximately infer the marginal distribution of a topic label assigned to a word called {\em message}.
At the M-step,
based on the normalized messages,
we estimate two multinomial parameters according to the maximum-likelihood criterion.
The EM algorithm iterates until converges to the local optimum.
On the other hand,
the NMF algorithm with KL divergence has a probabilistic interpretation~\citep{Lee:01},
which views the multiplication of two factorized matrices as the normalized probability distribution.
Notice that the widely-used performance measure perplexity~\citep{Blei:03,Asuncion:09,Hoffman:10} for topic models
follows exactly the same KL divergence in NMF,
which implies that the NMF algorithm may achieve a lower perplexity in learning topic models.
Therefore,
connecting NMF with LDA may inspire more efficient algorithms to learn topic models.
For example,
in this paper,
we show that the proposed TBP can avoid storing messages to reduce the memory usage.
More generally,
we speculate that finite mixture models and latent factor models~\citep{Gershman:12} may share similar learning techniques,
which may inspire more efficient training algorithms to each other in the near future.

\section{Experimental Results} \label{s4}

\begin{table}
\centering
\caption{Statistics of four document data sets.}
\begin{tabular}{|l|l|l|l|l|} \hline
Data sets &$D$       &$W$     &$N_d$      &$W_d$      \\ \hline \hline
ENRON     &$39861$   &$28102$ &$160.9$    &$93.1$     \\ \hline
NYTIMES   &$15000$   &$84258$ &$328.7$    &$230.2$    \\ \hline
PUBMED    &$80000$   &$76878$ &$68.4$     &$46.7$     \\ \hline
WIKI      &$10000$   &$77896$ &$1013.3$   &$447.2$     \\
\hline\end{tabular}
\label{dataset}
\end{table}

Our experiments aim to confirm the less memory usage of TBP compared with
the state-of-the-art batch learning algorithms such as VB~\citep{Blei:03}, GS~\citep{Griffiths:04} and BP~\citep{Zeng:11} algorithms,
as well as online learning algorithm such as online VB (OVB)~\citep{Hoffman:10}.
We use four publicly available document data sets~\citep{Porteous:08,Hoffman:10}:
ENRON, NYTIMES, PUBMED and WIKI.
Previous studies~\citep{Porteous:08} revealed that the topic modeling result is relatively insensitive to the total number of documents in the corpus.
Because of the memory limitation for GS, BP and VB algorithms,
we randomly select $15000$ documents from the original NYTIMES data set,
$80000$ documents from the original PUBMED data set,
and $10000$ documents from the original WIKI data set for experiments.
Table~\ref{dataset} summarizes the statistics of four data sets,
where $D$ is the total number of documents in the corpus,
$W$ is the number of words in the vocabulary,
$N_d$ is the average number of word tokens per document,
and $W_d$ is the average number of word indices per document.

We randomly partition each data set into halves with one for training set and the other for test set.
The training perplexity~\eqref{perplexity} is calculated on the training set in $500$ iterations.
Usually,
the training perplexity will decrease with the increase of number of training iterations.
The algorithm often converges if the change of training perplexity at successive iterations is less than a predefined threshold.
In our experiments,
we set the threshold as one because the decrease of training perplexity is very small after satisfying this threshold.
The predictive perplexity for the unseen test set is computed as follows~\citep{Asuncion:09}.
On the training set,
we estimate $\boldsymbol{\phi}$ from the same random initialization after $500$ iterations.
For the test set,
we randomly partition each document into $80\%$ and $20\%$ subsets.
Fixing $\boldsymbol{\phi}$,
we estimate $\boldsymbol{\theta}$ on the $80\%$ subset by training algorithms from the same random initialization after $500$ iterations,
and then calculate the predictive perplexity on the rest $20\%$ subset,
\begin{align}
\text{predictive perplexity}=\exp\Bigg\{-\frac{\sum_{w,d}
x_{w,d}^{20\%}\log(\boldsymbol{\phi}\boldsymbol{\theta})_{w,d}}
{\sum_{w,d} x_{w,d}^{20\%}}\Bigg\},
\end{align}
where $x_{w,d}^{20\%}$ denotes word counts in the the $20\%$ subset.
The lower predictive perplexity represents a better generalization ability.

\subsection{Comparison with Batch Learning Algorithms}

We compare TBP with other batch learning algorithms such as GS, BP and VB.
For all data sets,
we fix the same hyperparameters as
$\alpha = 2/K, \beta = 0.01$~\citep{Porteous:08}.
The CPU time per iteration is measured after sweeping the entire data set.
We report the average CPU time per iteration after $T = 500$ iterations,
which practically ensures that GS, BP and VB converge in terms of training perplexity.
For a fair comparison,
we use the same random initialization to examine all algorithms with $500$ iterations.
To repeat our experiments,
we have made all source codes and data sets publicly available~\citep{Zeng:12}.
These experiments are run on the Sun Fire X4270 M2 server with two 6-core $3.46$ GHz CPUs and $128$ GB RAMs.

\begin{table}[t]
\centering
\caption{Message memory (MBytes) for training set when $K=100$.}
\begin{tabular}{|c|c|c|c|c|} \hline
Inference methods   & ENRON     & NYTIMES   & PUBMED    & WIKI        \\ \hline \hline
VB and BP           & $1433.6$  & $1323.1$  & $1425.1$  & $1705.0$    \\ \hline
GS                  & $12.6$    & $9.5$     & $10.4$    & $19.3$      \\ \hline
aTBP and sTBP       & $0$       & $0$       & $0$       & $0$         \\ \hline
\end{tabular}
\label{memory}
\end{table}

Table~\ref{memory} compares the the message memory usage during training.
VB and BP consumes more than $1$GB memory to message passing when $K=100$.
VB and BP even require more than $9$GB for message passing when $K=900$,
because their message memory increases linearly with the number of topics $K$ in Eq.~\eqref{BPM}.
In contrast,
GS needs only $10\sim20$MB memory for message passing.
The advantage of GS is that its memory occupancy does not depend on the number of topics $K$ in Eq.~\eqref{GSM}.
Therefore,
PGS~\citep{Newman:09} can handle the relatively large-scale data set
containing thousands of topics without memory problems using the parallel architecture.
However,
PGS still requires message memory for message passing at the distributed computing unit.
Clearly,
both aTBP and sTBP do not need memory space to store previous messages,
and thus save a lot of memory usage.
This is a significant improvement especially compared with VB and BP algorithms.
In conclusion,
TBP is our first choice for batch topic modeling when memory is limited for topic modeling of massive corpora containing a large number of topics.

\begin{figure*}
\centering
\includegraphics[width=1\linewidth]{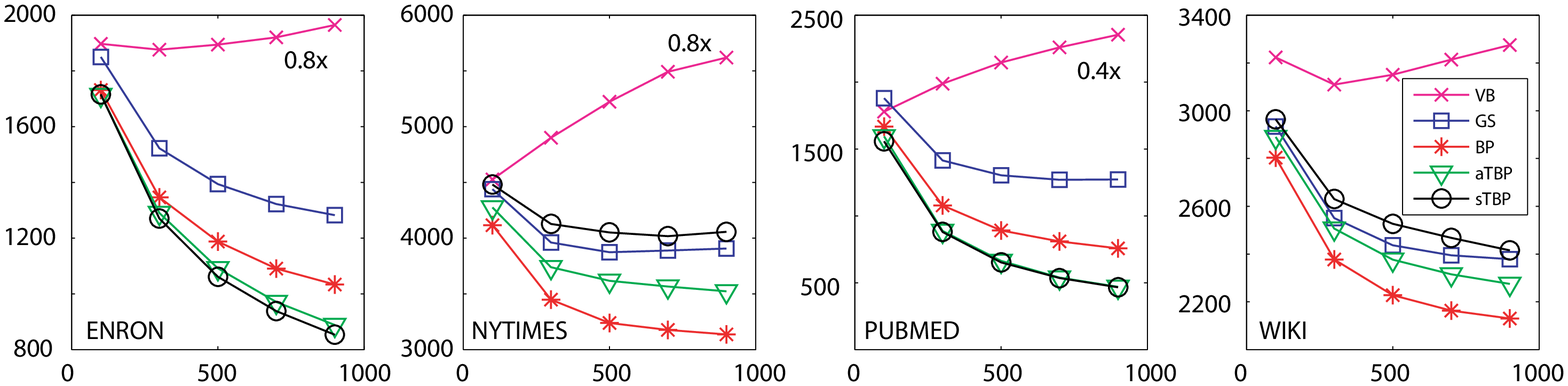}
\caption{Predictive perplexity as $K=\{100,300,500,700,900\}$ on ENRON, NYTIMES, PUBMED and WIKI data sets.
The notation $0.8x$ and $0.4x$ denote the predictive perplexity is multiplied by $0.8$ and $0.4$, respectively.}
\label{prediction}
\end{figure*}

Fig.~\ref{prediction} shows the topic modeling accuracy measured by the predictive perplexity on the unseen test set.
The lower predictive perplexity implies a better topic modeling performance.
Obviously,
VB performs the worst among all batch learning algorithms with the highest predictive perplexity.
For a better illustration,
we multiply VB's perplexity by $0.8$ on ENRON and NYTIMES, and by $0.4$ on PUBMED data sets, respectively.
Also,
we find that VB shows an overfitting phenomenon,
where the predictive perplexity increases with the increase of the number of topics $K$ on all data sets.
The basic reason is that VB optimizes an approximate variational distribution with the gap to the true distribution.
When the number of topics is large,
this gap cannot be ignored,
leading to serious biases.
We see that GS performs much better than VB on all data sets,
because it theoretically approximates the true distribution by sampling techniques.
BP always achieves a much lower predictive perplexity than GS,
because it retains all uncertainty of messages without sampling.
Both sTBP and aTBP perform equally well on ENRON and PUBMED data sets,
which also achieve the lowest predictive perplexity among all batch training algorithms.
However,
BP outperforms both sTBP and aTBP on NYTIMES and WIKI data sets.
Also,
aTBP outperforms both sTBP and GS,
while sTBP performs slightly worse than GS.
Because aTBP has consistently better topic modeling accuracy than GS on all data sets,
we advocate aTBP for topic modeling in limited memory.
As we discussed in Section~\ref{3.1},
BP/TBP has the lowest predictive perplexity mainly because it directly minimizes the KL divergence
between $\mathbf{x}$ and $\boldsymbol{\phi}\boldsymbol{\theta}$ from the NMF perspective.

\begin{figure*}
\centering
\includegraphics[width=1\linewidth]{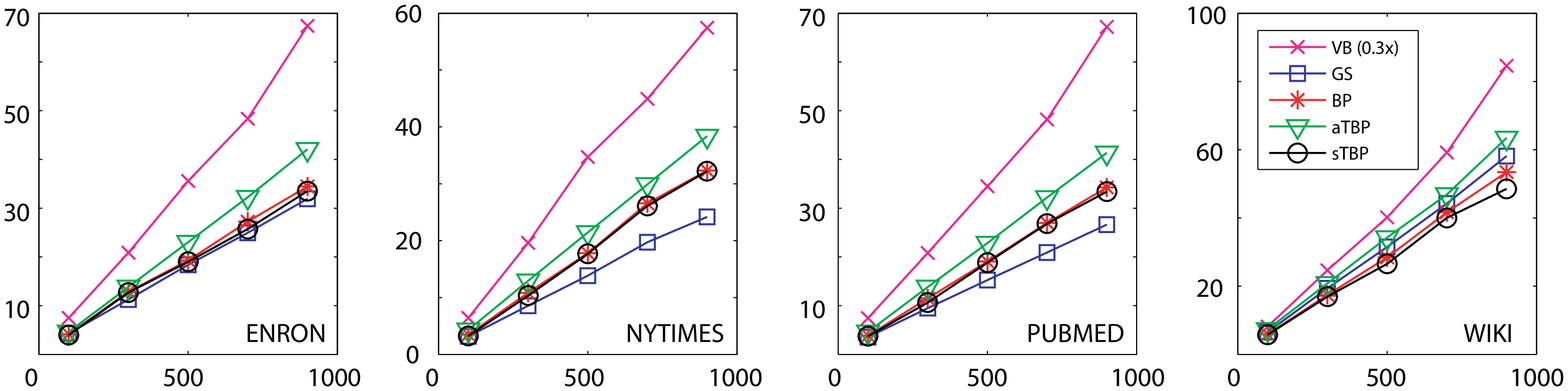}
\caption{CPU time per iteration (seconds) as $K=\{100,300,500,700,900\}$ on ENRON, NYTIMES, PUBMED and WIKI data sets.
The notation $0.3x$ denotes the training time is multiplied by $0.3$.}
\label{time}
\end{figure*}

Fig.~\ref{time} shows the CPU time per iteration of all algorithms.
All these algorithms has a linear time complexity of $K$.
VB is the most time-consuming because it involves complicated digamma function computation~\citep{Asuncion:09,Zeng:11}.
For a better illustration,
we multiply the VB's training time by $0.3$.
Although BP runs faster than GS when $K$ is small ($K \le 100$)~\citep{Zeng:11},
it is sometimes slower than GS when $K$ is large ($K \ge 100$), especially on ENRON and PUBMED data sets.
The reason lies in that GS often randomly samples a topic label without visiting all $K$ topics,
while BP requires searching all $K$ topics for the message update.
When $K$ is very large,
this slight difference will be enlarged.
sTBP runs as fast as BP in most cases,
but aTBP runs slightly slower than both sTBP and BP.
Comparing two algorithms in Figs.~\ref{stbp} and~\ref{atbp},
we find that aTBP uses more division operations than sTBP at each training iteration,
which accounts for aTBP's slowness.
As a summary,
TBP has a comparable topic modeling speed as GS and BP but with reduced memory usage.

\begin{figure*}
\centering
\includegraphics[width=1\linewidth]{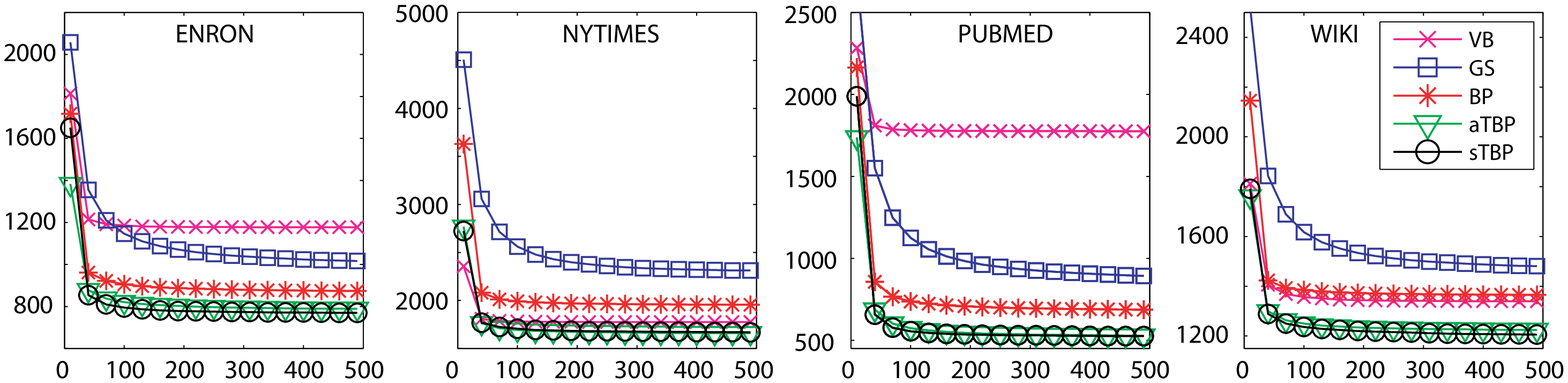}
\caption{Training perplexity as a function of the number of iterations when $K=500$ on ENRON, NYTIMES, PUBMED and WIKI data sets.}
\label{convergence}
\end{figure*}

Fig.~\ref{convergence} shows the training perplexity as a function of training iterations.
All algorithms converge to a fixed point given enough training iterations.
On all data sets,
VB usually uses $110\sim170$ iterations,
GS uses around $400\sim470$ iterations,
and BP/TBP uses $180\sim230$ iterations for convergence.
Although the digamma function calculation is slow,
it reduces the number of training iterations of VB to reach convergence.
GS is a stochastic sampling method,
and thus requires more iterations to approximate the true distribution.
Because BP/TBP is a deterministic message passing method,
it needs less iterations to achieve convergence than GS.
Overall,
BP/TBP consumes the least training time until convergence according to Figs.~\ref{time} and~\ref{convergence}.

\begin{figure*}
\centering
\includegraphics[width=1\linewidth]{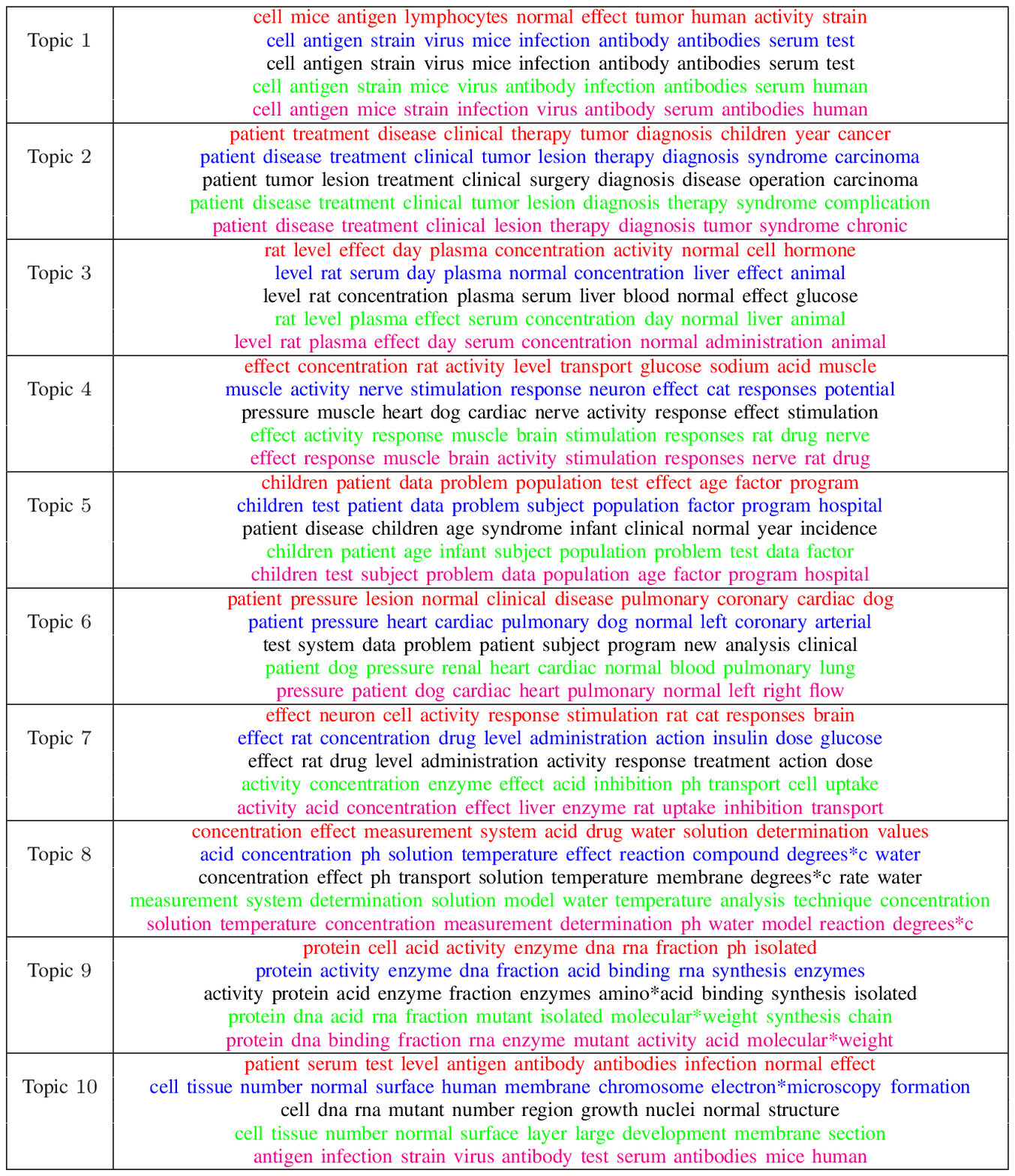}
\caption{Top ten words in ten topics extracted from the subset of PUBMED: VB (red), GS (blue), BP (black), aTBP (green) and sTBP (magenta).
Most topics contain similar words but with a different order.
The subjective measures~\citep{Chang:09b} such as word intrusions in topics and topic intrusions in documents are comparable
among different algorithms.}
\label{topic}
\end{figure*}

Fig.~\ref{topic} shows the top ten words of $K=10$ topics extracted by VB (red), GS (blue), BP (black), aTBP (green) and sTBP (magenta).
We see that most topics contain similar top ten words but with a different order.
More formally,
we can adopt subjective measures such as the word intrusion in topics and the topic intrusion in documents~\citep{Chang:09b}
to evaluate extracted topics.
PUBMED is a biomedical corpus.
According to our prior knowledge in biomedical domain,
we find these topics are all meaningful.
Under this condition,
we advocate TBP for topic modeling with reduced memory requirements.

\subsection{Comparison with Online Algorithms}

\begin{figure*}
\centering
\includegraphics[width=0.8\linewidth]{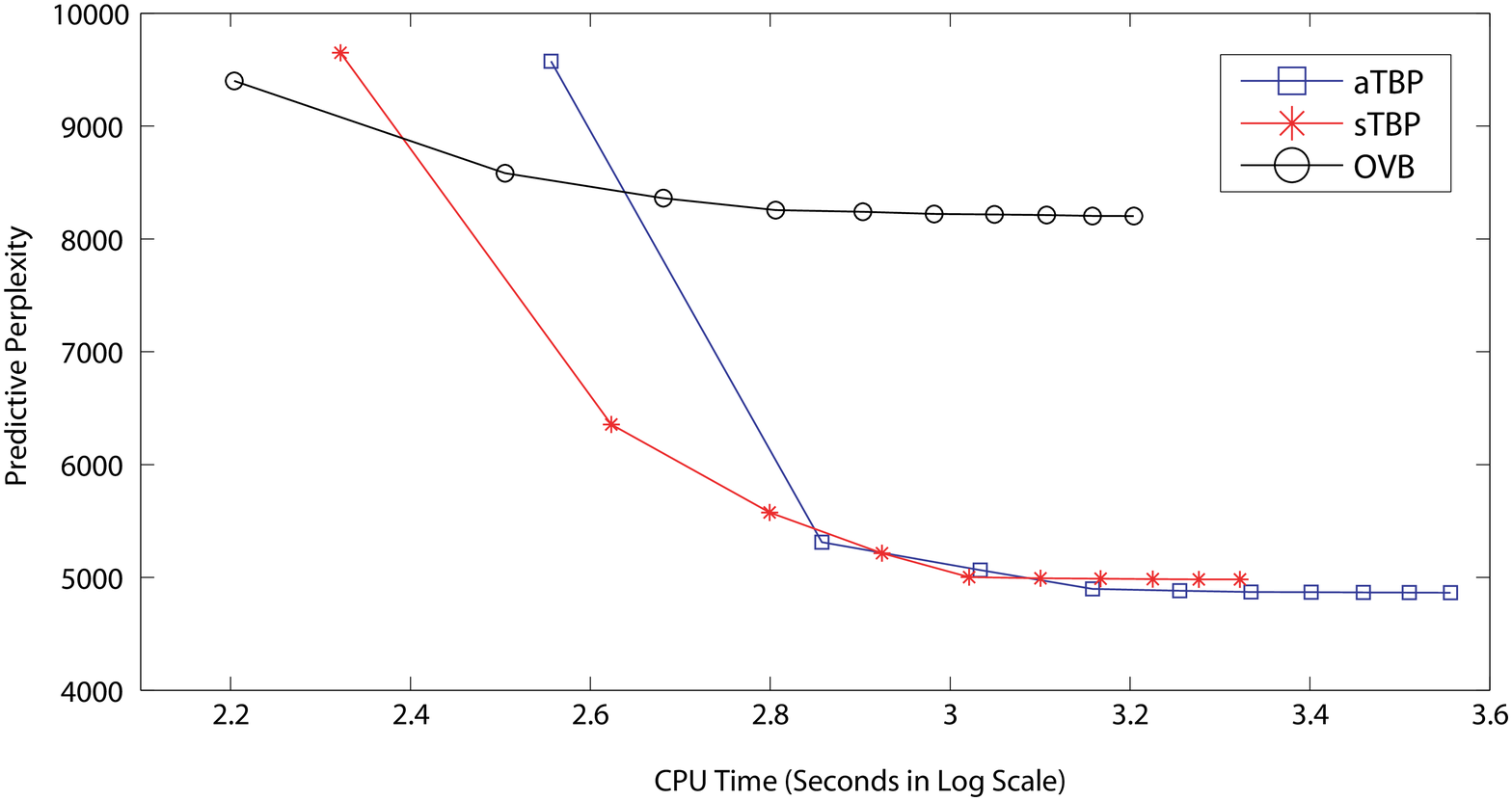}
\caption{Predictive perplexity obtained on the complete PUBMED corpus as a function of CPU time (seconds in log scale) when $K=10$.}
\label{online}
\end{figure*}

We compare the topic modeling performance between TBP and the state-of-the-art online topic modeling algorithm
OVB~\citep{Hoffman:10}\footnote{\url{http://www.cs.princeton.edu/~blei/topicmodeling.html}}
on a desktop computer with $2$GB memory.
The complete $7$GB PUBMED data set~\citep{Porteous:08}
contains a total of $D = 820,000,000$ documents with a vocabulary size $W = 141,043$.
Currently,
only TBP and online topic modeling methods can handle $7$GB data set using $2$GB memory.
OVB~\citep{Hoffman:10} uses the following default parameters:
$\kappa = 0.5$, $\tau_0 = 1024$, and the mini-batch size $S = 1024$.
We randomly reserve $40,000$ documents as the test set,
and use the remainder $8,160,000$ documents as the training set.
The number of topics $K=10$.
The hyperparameters $\alpha = 2/K = 0.05$ and $\beta = 0.01$.

Fig.~\ref{online} shows the predictive perplexity as a function of training time (seconds in log scale).
OVB converges slower than TBP because it reads input data as a data stream,
discarding each mini-batch sequentially after one look.
Notice that,
for each mini-batch,
OVB still requires allocating message memory for computation.
In contrast,
TBP achieves a much lower perplexity using less memory usage and training time.
There are two major reasons.
First,
TBP directly optimizes the perplexity in terms of the KL divergence in Eq~\eqref{perplexity},
so that it can achieve a much lower perplexity than OVB.
Second,
OVB involves computationally expensive digamma functions which significantly slow down the speed.
We see that sTBP is a bit faster than aTBP because it does not perform the division operation at each iteration (see Figs.~\ref{stbp} and~\ref{atbp}).
Because aTBP influences matrix factorization immediately after the matrix update,
it converges at a slightly lower perplexity than sTBP.

\section{Conclusions} \label{s5}

This paper has presented a novel tiny belief propagation (TBP) algorithm for training LDA with significantly reduced memory requirements.
The TBP algorithm reduces the message memory required by conventional message passing algorithms including GS, BP and VB.
We also discuss the intrinsic relation between the proposed TBP and NMF~\citep{Lee:01} with KL divergence.
We find that TBP can be approximately viewed as a special NMF algorithm for minimizing the perplexity metric,
which is a widely-used evaluation method for different training algorithms of LDA.
In addition,
we confirm the superior topic modeling accuracy of TBP in terms of predictive perplexity on extensive experiments.
For example,
when compared with the state-of-the-art online topic modeling algorithm OVB,
the proposed TBP is faster and more accurate to extract $10$ topics from $7$GB PUBMED corpus using a desktop computer with $2$GB memory.

Recently,
the NMF algorithm has been advocated to learn topic models such as LDA with a polynomial time~\citep{Arora:12}.
The proposed TBP algorithm also suggests that the NMF algorithms can be applied to
training topic models like LDA with a high accuracy in terms of the perplexity metric.
We hope that our results may inspire more and more NMF algorithms~\citep{Lee:01} to be extended to
learn other complicated LDA-based topic models~\citep{Blei:12} in the near future.

\acks{This work is supported by NSFC (Grant No. 61003154),
the Shanghai Key Laboratory of Intelligent Information Processing,
China (Grant No. IIPL-2010-009),
and a grant from Baidu to JZ,
and a GRF grant from RGC UGC Hong Kong (GRF Project No.9041574)
and a grant from City University of Hong Kong (Project No. 7008026) to ZQL.}


%

\vskip 0.2in
\bibliographystyle{natbib}
\bibliography{IEEEabrv,TBP}

\end{document}